\documentclass{bmvc2k}

\usepackage{preamble} % NOTE: Ezt Iván csinálta hogy tömörebb legyen a cikk forrása. (lásd preamble.sty)

\title{Optimal Multi-view Correction of Local Affine Frames}

\addauthor{Ivan Eichhardt}{ivan.eichhardt@sztaki.mta.hu}{1}
\addauthor{Daniel Barath}{barath.daniel@sztaki.mta.hu}{2}

\addinstitution{
 Machine Perception Research Laboratory,
 MTA SZTAKI, Budapest, Hungary
}
\addinstitution{
 Centre for Machine Perception, Department of Cybernetics
 Czech Technical University, Prague, Czech Republic
}

\runninghead{I.\ Eichhardt, D.\ Barath}{Multi-view Correction of Affine Frames}

\begin{document}

\maketitle

%%%%%%%%% ABSTRACT
\begin{abstract}
    The technique requires the epipolar geometry to be pre-estimated between each image pair. 
    It exploits the constraints which the camera movement implies, in order to apply a closed-form correction to the parameters of the input affinities. 
    Also, it is shown that the rotations and scales obtained by partially affine-covariant detectors, \eg AKAZE or SIFT, can be completed to be full affine frames by the proposed algorithm. 
    It is validated both in synthetic experiments and on publicly available real-world datasets that the method always improves the output of the evaluated affine-covariant feature detectors. As a by-product, these detectors are compared and the ones obtaining the most accurate affine frames are reported. 
    For demonstrating the applicability, we show that the proposed technique as a pre-processing step improves the accuracy of pose estimation for a camera rig, surface normal and homography estimation. 
\end{abstract}

%%%%%%%%% BODY TEXT
%%%%%%%%%%%%%%%%%%%%%%%%%%%%%%%%%%%%%%%%%%%%%%%%%%%%%%%%%%%%%%%%%%%%%%%%%%%%%%%%
%%%%%%%%%%%%%%%%%%%%%%%%%%%%%%%%%%%%%%%%%%%%%%%%%%%%%%%%%%%%%%%%%%%%%%%%%%%%%%%%
%%%%%%%%%%%%%%%%%%%%%%%%%%%%%%%%%%%%%%%%%%%%%%%%%%%%%%%%%%%%%%%%%%%%%%%%%%%%%%%%
\section{Introduction}

A method is proposed for estimating local affine frames (LAFs) accurately in a rigid\footnote{The generalisation to multiple rigid motions each satisfying a different constraint is straightforward.} scene observed by multiple cameras. 
In particular, we are interested in finding the affine mappings which are the closest in the least squares sense to the detected ones and, also, for which the constraints implied by the camera movement hold. 
The method takes a sequence of affine features detected by an affine-covariant feature detector, \eg, Affine-SIFT~\cite{Morel2009}, and returns the affinities corrected by the proposed closed-form procedure.
Also, the method is applicable when a not fully affine-covariant detector is used, \eg AKAZE~\cite{detectors_akaze_alcantarilla_fast_2013} or SIFT~\cite{SIFT2004} for which LAFs are only partially estimated, \eg scales and orientations.
The proposed method returns the underlying affine frames consistent with the camera movement.% in this case, too. %and interpreting the given affine components the most. 

\begin{figure}[t]
    \centering
    	\subfigure[]{
            \includegraphics[width=0.4\linewidth]{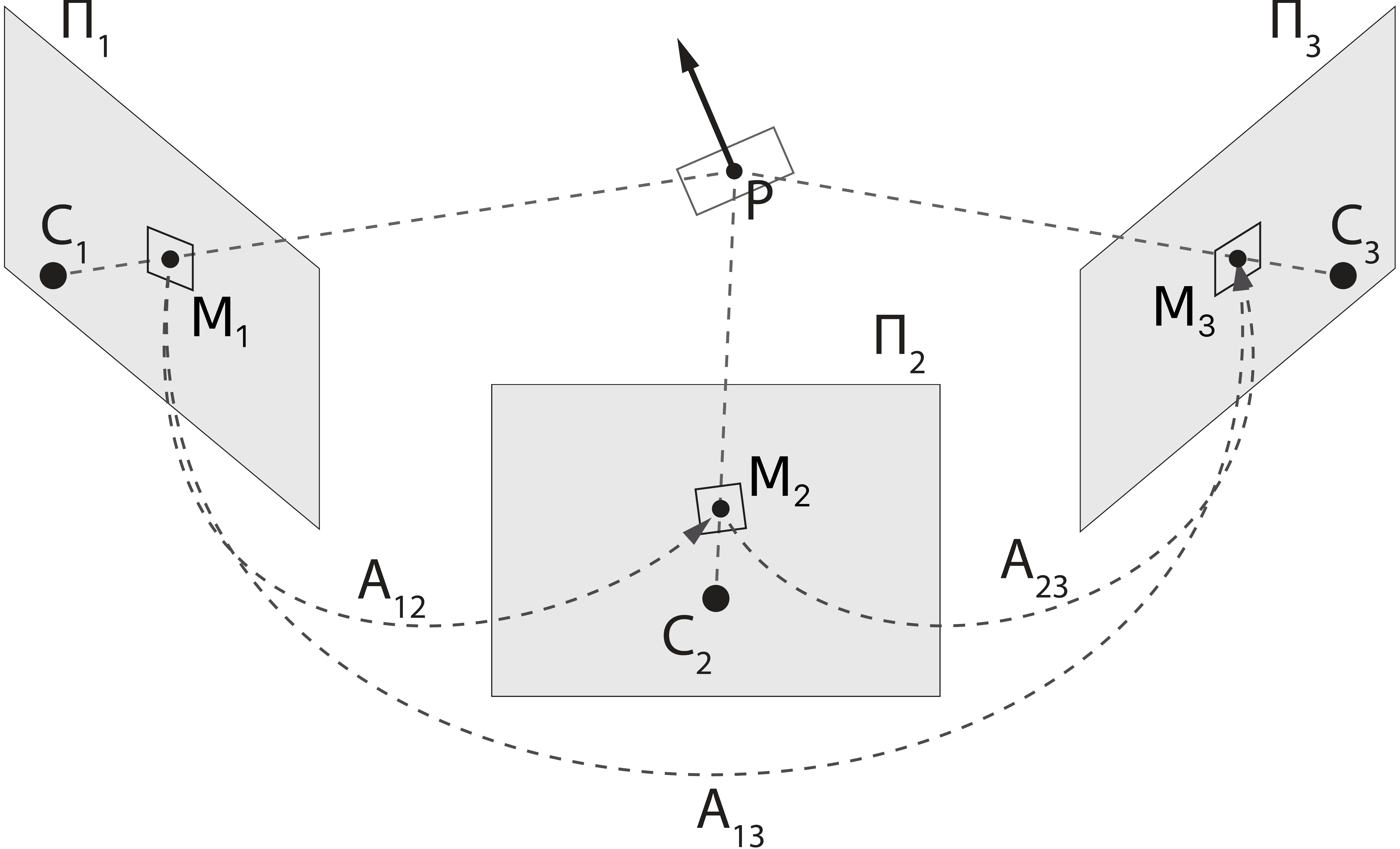}
    	}
    	\subfigure[]{
            \includegraphics[width=0.55\linewidth]{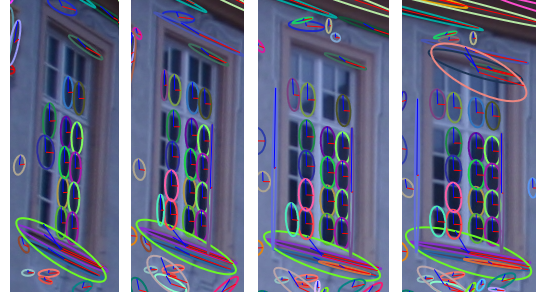}
    	}%\hfill
    \caption{\textbf{(a)} Three cameras ($\mat{C}_1$, $\mat{C}_2$, $\mat{C}_3$) observing point $\textbf{P}$. 
    The shape of the region induced by the plane on which $\textbf{P}$ lies in the $i$th image is described by local affine frame $\matM_i$ (LAF). 
    The LAFs around the projected points between the $i$th and $j$th views are related by local affine transformation $\textbf{A}_{ij}$. \textbf{(b)} Example multi-view region correspondences represented by oriented ellipses (LAFs) across multiple views. Corresponding ellipses are denoted by colour. 
    %Corresponding image regions across multiple views. The ellipses represent the linear transformation part of detected LAFs. Matching LAFs share the same color.
    }
    \label{fig:corresponding_local_affine_frames}
\end{figure}

Nowadays, a number of algorithms have been proposed for solving various computer vision problems by exploiting the geometric information provided by affine correspondences. 
For instance, Perdoch \etal~\cite{perd2006epipolar} proposed techniques for approximating the epipolar geometry between two images by generating point correspondences from the affine features. 
Bentolila and Francos~\cite{bentolila2014conic} showed a method to estimate the exact, \ie, with no approximation, fundamental matrix using three correspondences. Raposo \etal~\cite{raposo2016theory} proposed a solution for essential matrix estimation using two feature pairs. Eichhardt and Chetverikov~\cite{eichhardtChetverikovECCV2018} proposed a generalisation of the approach considering arbitrary central projection. Bar\'ath \etal~\cite{barath2017focal} proved that even the semi-calibrated case, \ie, when the objective is to find the essential matrix and a common focal length, is solvable from two correspondences. 
Homographies can also be estimated from two features~\cite{koser2009geometric} without any a priori knowledge about the camera movement. In case of known epipolar geometry, a single affine correspondence is sufficient for estimating a homography~\cite{barathPRL2017}. 
Affine correspondences were successfully used in multi-homography estimation~\cite{barath2017multi}.
Also, affine frames encode the surface normals~\cite{Molnar2014}. Therefore, if the cameras are calibrated, the normal can be unambiguously estimated from a single correspondence~\cite{koser2009geometric}. 
Multiple-view normal estimation~\cite{barathEichhardtHajderTIP2019,eichhardtHajderICCVW2017} is also possible.
Pritts \etal~\cite{Pritts2017RadiallyDistortedCT,Pritts2018RadiallyDistortedScales} showed that the radial distortion parameters can be retrieved, as well. 

Affine correspondences encode higher-order information about the underlying scene geometry. 
This is what makes the listed algorithms able to estimate geometric models, \eg, homographies and fundamental matrices, using significantly fewer correspondences than point-based methods.
Being more complex than 2D points, the accurate estimation of affine frames is a more complicated task. 
The estimation is, in practice, done by applying an affine- or partially affine-covariant feature detector which simultaneously recovers points and the corresponding affine frames. Some methods investigate the shapes of corresponding image regions (\eg, MSER~\cite{MSER}, TMBR~\cite{detectors_tbmr_xu2014tree}). 
Other techniques generate synthetic views by transforming the input images by affine transformations (\eg, ASIFT~\cite{Morel2009}, MODS~\cite{mishkin2015mods}), whilst some of them optimise each detected feature by minimising a photo-consistency-based cost function~\cite{mikolajczyk2005comparison}. 
However, affine correspondences are significantly more noisy than points even when applying state-of-the-art feature detectors.
%(see Figure). (NOTE: put some table/figure showing it)

Barath \etal~\cite{barath2016accurate} proposed two geometric constraints describing the relationship of stereo epipolar geometry and affine correspondences.
The constraints are built on the fact that a geometrically valid affine frame must transform the normals of the corresponding epipolar lines into each other. Also, the scaling factor along the normal direction is determined by the epipolar geometry and, thus, can be calculated from the fundamental matrix. 
Exploiting these constraints, the EG-$L_2$-Optimal algorithm is proposed in~\cite{barath2016accurate} to make an input affine correspondence consistent with the fundamental matrix by an efficient closed-form approach.

In this paper, we extend the EG-$L_2$-Optimal technique by generalising the constraints to multiple views. 
The proposed method is applicable when a sequence of corresponding affine frames is given through multiple images (see Fig.~\ref{fig:corresponding_local_affine_frames}). 
It is efficient due to being solved by a closed-form approach.
It is validated both on synthetic experiments and on a number of real-world datasets that the method always improves the output of state-of-the-art affine- and partially affine-covariant feature detectors. 
As a by-product, these detectors are compared and the best ones, in terms of finding the most geometrically accurate affine frames, are reported.
As possible applications, it is shown that the proposed method improves homography and surface normal estimation. 
Also, using the corrected affine frames makes the relative motion estimation of a camera rig more accurate. 

%%%%%%%%%%%%%%%%%%%%%%%%%%%%%%%%%%%%%%%%%%%%%%%%%%%%%%%%%%%%%%%%%%%%%%%%%%%%%%%%
%%%%%%%%%%%%%%%%%%%%%%%%%%%%%%%%%%%%%%%%%%%%%%%%%%%%%%%%%%%%%%%%%%%%%%%%%%%%%%%%
%%%%%%%%%%%%%%%%%%%%%%%%%%%%%%%%%%%%%%%%%%%%%%%%%%%%%%%%%%%%%%%%%%%%%%%%%%%%%%%%
\section{Epipolar constraints on affine features}

In this section, first, the required theoretical background is discussed. Then we show the constraints which a pair of affine frames imply on the two-view epipolar geometry. 

%%%%%%%%%%%%%%%%%%%%%%%%%%%%%%%%%%%%%%%%%%%%%%%%%%%%%%%%%%%%%%%%%%%%%%%%%%%%%%%%
%%%%%%%%%%%%%%%%%%%%%%%%%%%%%%%%%%%%%%%%%%%%%%%%%%%%%%%%%%%%%%%%%%%%%%%%%%%%%%%%
\paragraph{Notation and preliminaries.}

A \textit{local affine frame} (LAF) is a pair $\brc{\pt{x}, \matM}$ of a point $\pt{x} = [u, \; v]^\trans$ and a $2 \times 2$ linear transformation $\matM \in \mathbb{R}^{2 \times 2}$. 
Matrix $\matM$ is defined by the partial derivatives, \wrt the image directions, of the projection function~\cite{Barath2015}. 
An \textit{affine correspondence}
$(\pt{x}_1, \pt{x}_2, \matA)$ is a triplet, where $\pt{x}_1 = [u_1 \; v_1]^\trans$ and $\pt{x}_2 = [u_2 \; v_2]^\trans$ is a corresponding pair of points in two images and $\matA$
is a $2 \times 2$ linear transformation which is called \textit{local affine transformation} and defined as $\matA = \matM_2^{} \matM_1^{-1}$, where $\matM_i$ is the matrix from the corresponding LAF in the $i$th image, $i \in \{1,2\}$. 
%Its elements in a row-major order are: $a_1$, $a_2$, $a_3$, and $a_4$. 

%The $j$th element of the \textbf{fundamental} ($\mat{F}$) and essential matrices ($\mat{E}$) in row-major order is denoted as $f_j$ and $e_j$, respectively, $j \in [1,9]$.
%
The fundamental ($\mat{F}$) and essential ($\mat{E}$) matrices ensure the epipolar constraint as $\tilde{\pt{x}}_2^\trans \mat{F} \tilde{\pt{x}}_1 = \tilde{\pt{x}}_2^\trans \mat{K}_2^{-\trans} \mat{E} \mat{K}_1^{-1} \tilde{\pt{x}}_1 = 0$, where $\mat{K}_i$ is the intrinsic calibration matrix of the $i$th camera and $\tilde{\pt{x}}_i$ 
%= \begin{bmatrix}\pt{x}_i^\trans &  1\end{bmatrix}^\trans$ 
is the homogeneous form of point $\pt{x}_i$.

%%%%%%%%%%%%%%%%%%%%%%%%%%%%%%%%%%%%%%%%%%%%%%%%%%%%%%%%%%%%%%%%%%%%%%%%%%%%%%%%
%%%%%%%%%%%%%%%%%%%%%%%%%%%%%%%%%%%%%%%%%%%%%%%%%%%%%%%%%%%%%%%%%%%%%%%%%%%%%%%%
%\subsection{Constraints on affine-covariant features}

\paragraph{Constraints on affine correspondences.}
Suppose that we are given an affine correspondence $\brc{\pt{x}_1, \pt{x}_2, \matA}$ constructed from two LAFs $\brc{\pt{x}_1, \matM_1}$ and $\brc{\pt{x}_2, \matM_2}$ such that 
\begin{equation}
    \matA = \matM_2^{} \matM_1^{-1}.
    \label{eq:affine-laf-relation}
\end{equation}
In case of pinhole cameras, the following constraint~\cite{eichhardtChetverikovECCV2018} holds.
\begin{equation}
    \matA^\trans \underbrace{\mat{I}_{2\times3}\mat{F}\tilde{\pt{x}}_1}_{\vect{a}} + \underbrace{\mat{I}_{2\times3}\mat{F}^\trans\tilde{\pt{x}}_2}_{\vect{b}} = \mathbf{0},
    \label{eq:fundamental-epipolar-constraint}
\end{equation}
where $\mat{I}_{2 \times 3}$ is a $2 \times 3$ identity matrix and $\mat{F}$ is the fundamental matrix.
Note that, in case of arbitrary central projection, $\mat{a} = \nabla q_2^\trans\mat{E}q_1$ and $\mat{b} = \nabla q_1^\trans\mat{E}^\trans q_2$, where $q_i$ is the bearing vector corresponding to $\pt{x}_i$ and $\nabla q_i$ is its gradient \wrt $\pt{x}_i$. This relationship is described in \cite{eichhardtChetverikovECCV2018} in depth.
A compact form of the expression is $\matA^\trans \vect{a} + \vect{b} = \mathbf{0}$.

%%%%%%%%%%%%%%%%%%%%%%%%%%%%%%%%%%%%%%%%%%%%%%%%%%%%%%%%%%%%%%%%%%%%%%%%%%%%%%%%
%%%%%%%%%%%%%%%%%%%%%%%%%%%%%%%%%%%%%%%%%%%%%%%%%%%%%%%%%%%%%%%%%%%%%%%%%%%%%%%%
%%%%%%%%%%%%%%%%%%%%%%%%%%%%%%%%%%%%%%%%%%%%%%%%%%%%%%%%%%%%%%%%%%%%%%%%%%%%%%%%
\paragraph{Constraints on local affine frames.}
In order to define how a pair of LAFs is constrained by the epipolar geometry, we plug formula \eqref{eq:affine-laf-relation} into \eqref{eq:fundamental-epipolar-constraint}. The obtained equation is as follows: 
%
%\begin{equation}
%\begin{aligned}
    $\matA^\trans \vect{a} + \vect{b} = \brc{\matM_2^{} \matM_1^{-1}}^\trans \vect{a} + \vect{b} =
    \matM_1^{-\trans}\matM_2^\trans \vect{a} + \vect{b} = \vect{0}$.
%\end{aligned}
%\end{equation}
%
After left-multiplying the expression by $\matM^{-\trans}$, the following epipolar constraint on a pair of LAFs is given:
\begin{equation} 
    \matM_2^\trans \vect{a} + \matM_1^\trans \vect{b} = \mathbf{0}.
    \label{eq:epipolar_constraint_on_corresponding_LAF}
\end{equation}

%%%%%%%%%%%%%%%%%%%%%%%%%%%%%%%%%%%%%%%%%%%%%%%%%%%%%%%%%%%%%%%%%%%%%%%%%%%%%%%%
%%%%%%%%%%%%%%%%%%%%%%%%%%%%%%%%%%%%%%%%%%%%%%%%%%%%%%%%%%%%%%%%%%%%%%%%%%%%%%%%
%%%%%%%%%%%%%%%%%%%%%%%%%%%%%%%%%%%%%%%%%%%%%%%%%%%%%%%%%%%%%%%%%%%%%%%%%%%%%%%%
\section{Multi-view EG-$L_2$-Optimal correction}

Let $\mathcal{V}$ be the set of views in a multiple-view correspondence, \ie $\pt{x}_k$ ($\forall k\in\mathcal{V}$) are projections of the same point in space where $\brc{\pt{x}_k, \hat{\matM}_k}$ is the respective LAF. % detected by, \eg, an affine-covariant feature detector. 
 The set of pairwise correspondences is $\mathcal{C}\subseteq\mathcal{V}\times\mathcal{V}$.
The objective is to find all $\matM_k$, such that
\begin{equation}
\begin{aligned}
    \min_{\matM_k} & \sum_{k\in\mathcal{V}} \brcnorm{ {\matM}_k^\trans - \hat{\matM}_k^\trans }_\text{F}^{2} & \text{\quad s.t. \quad} & \forall \brc{i,j} \in \mathcal{C} \,:\, \matM_j^\trans \vect{a}_{ij} + \matM_i^\trans \vect{b}_{ij} = \mathbf{0},
\end{aligned}
\label{eq:problem-to-optimize}
\end{equation}
where $\vect{a}_{ij}$ and $\vect{b}_{ij}$ are as defined above, \eg $\vect{a}_{ij}=\nabla q_j^\trans\mat{E}q_i$ and $\vect{b}_{ij}=\nabla q_i^\trans\mat{E}^\trans q_j$ for the pair $\brc{i,j}$ of views.
An equivalent form of \eqref{eq:problem-to-optimize} using Lagrange multipliers $\vect{\lambda}_{ij}\in\mathbb{R}^{2}$ is as follows: %for $\brc{i,j} \in \mathcal{C}$
\begin{equation}
\begin{aligned}
    \min_{\matM_k,\vect{\lambda}_{ij}} & \sum_{k\in\mathcal{V}} \frac{1}{2}\brcnorm{ {\matM}_k^\trans - \hat{\matM}_k^\trans }_\text{F}^{2}  -
    \sum_{\brc{i,j} \in \mathcal{C}} \vect{\lambda}_{ij}^\trans \brc{\matM_j^\trans \vect{a}_{ij} + \matM_i^\trans \vect{b}_{ij}}.
\end{aligned}
\label{eq:lagrange-opt}
\end{equation}
%
%where $\lambda_{ij} \in \mathbb{R}^2$.

\paragraph{Optimality conditions.}
To find the globally optimal solution, the 1st-order optimality conditions have to be investigated.
For each $k\in\mathcal{V}$, the gradient $\nabla_{\matM_k^\trans}$ of the expression in \eqref{eq:lagrange-opt} is % as follows:
\begin{equation}
    \matM_k^\trans  - \sum_{\brc{i,k} \in \mathcal{C}} \vect{\lambda}_{ik}\vect{a}_{ik}^\trans - \sum_{\brc{k,j} \in \mathcal{C}} \vect{\lambda}_{kj}\vect{b}_{kj}^\trans= \hat{\matM}_k^\trans,
    \label{eq:optimality_1}
\end{equation}
The gradient $\nabla_{\vect{\lambda}_{mn}}$ of \eqref{eq:lagrange-opt} corresponding to the Lagrange multiplier $\vect{\lambda}_{mn}$ gives an expression resembling the epipolar constraints in \eqref{eq:epipolar_constraint_on_corresponding_LAF}:
\begin{equation}
    \matM_n^\trans \vect{a}_{mn} + \matM_m^\trans \vect{b}_{mn} = \mathbf{0}.
    \label{eq:optimality_2}
\end{equation}
Given all the 1st-order optimality conditions, an equivalent form can be constructed as a single linear system as follows: 
\begin{equation}
    \begin{bmatrix}\mat{I}_{2\brcnorma{\mathcal{V}}\times2\brcnorma{\mathcal{V}}} & \mat{B} \\ \mat{B}^\trans & \mat{0}_{\brcnorma{\mathcal{C}}\times\brcnorma{\mathcal{C}}}\end{bmatrix}
    \begin{bmatrix}\matOmega \\ \matLambda \end{bmatrix}
    =
     \begin{bmatrix}\hat{\matOmega}  \\ \mat{0}_{\brcnorma{\mathcal{C}}\times2} \end{bmatrix},
\label{eq:matrixform}
\end{equation}
where $\matOmega=\begin{bmatrix}\mat{M}_1^\trans & \dots & \mat{M}_{\brcnorma{\mathcal{V}}}^\trans\end{bmatrix}^\trans$, $\hat{\matOmega}=\begin{bmatrix}\hat{\mat{M}}_1^\trans & \dots & \hat{\mat{M}}_{\brcnorma{\mathcal{V}}}^\trans\end{bmatrix}^\trans$ and $\matLambda=\begin{bmatrix}\dots & \lambda_{ij} & \dots \end{bmatrix}^\trans.$
%
%\begin{equation*}
%    \begin{aligned}
%        \matOmega=&\begin{bmatrix}\mat{M}_1^\trans & \dots & %\mat{M}_{\brcnorma{\mathcal{V}}}^\trans\end{bmatrix}^\trans, \quad
%        \hat{\matOmega}=&\begin{bmatrix}\hat{\mat{M}}_1^\trans & \dots & %\hat{\mat{M}}_{\brcnorma{\mathcal{V}}}^\trans\end{bmatrix}^\trans, \quad
%        \matLambda=&\begin{bmatrix}\dots & \lambda_{ij} & \dots %\end{bmatrix}^\trans.
%    \end{aligned}
%\end{equation*}

Note that $\begin{bmatrix}\mat{I}_{2\brcnorma{\mathcal{V}}\times2\brcnorma{\mathcal{V}}} & \mat{B}\end{bmatrix}$ encodes the optimality conditions in \eqref{eq:optimality_1}, and $\mat{B}^\trans\in\mathbb{R}^{\brcnorma{\mathcal{C}}\times2\brcnorma{\mathcal{V}}}$ holds the optimality conditions of \eqref{eq:optimality_2}. 
%
%A schematic depiction of a possible structure of $\mat{B}$ is as follows:
%%
%\begin{equation}
%\begin{aligned}
%\mat{B} =& \begin{bmatrix}\vect{b}_{12}^\trans & %\vect{a}_{12}^\trans&\vect{0}^\trans&\dots&\vect{0}^\trans\\&&\vdots&&\\\dots & \vect{b}_{ij}^\trans & \dots & %\vect{a}_{ij}^\trans & \dots\\&&\vdots&&\end{bmatrix}^\trans.
%\end{aligned}
%\end{equation}
%
Each line of $\mat{B}^\trans$ holds $\vect{a}_{ij}$ and $\vect{b}_{ij}$ needed for an epipolar constraint:
$\mat{B}^\trans\matOmega$ is zero, if $\matOmega$ stores LAFs consistent with the epipolar geometry.

%%%%%%%%%%%%%%%%%%%%%%%%%%%%%%%%%%%%%%%%%%%%%%%%%%%%%%%%%%%%%%%%%%%%%%%%%%%%%%%%
\paragraph{Efficient solution to the linear system.} Thanks to the the block matrix structure of \eqref{eq:matrixform}, formula 
%\begin{equation}
    $\matOmega
    =
    \hat{\matOmega} - \mat{B} \brc{\mat{B}^\trans\mat{B}}^{-1} \mat{B}^\trans \hat{\matOmega}$,
%\end{equation}
%
can be used to compute the optimal solution, where $\mat{B} \brc{\mat{B}^\trans\mat{B}}^{-1} \mat{B}^\trans$ is a projection matrix into the column space of $\mat{B}$. 
%Note that $\brc{\mat{B}^\trans\mat{B}}^{-1} \mat{B}^\trans$ is also the pseudo-inverse of $\mat{B}$, \ie, an equivalent solution is $\matOmega=\hat{\matOmega} - \mat{B}\mat{B}^\text{+}\hat{\matOmega}$.
%
%However, to avoid numerical instability, neither the direct computation of the inverse nor the pseudo-inverse is preferred.
To avoid numerical instability, the direct computation of the inverse is not preferred.
To our experiments, the most stable solution is given by the column-pivoting Householder QR decomposition of \mat{B}, in case \mat{B} is noise-free.
In a Structure-from-Motion (SfM) system, it can be guaranteed that \mat{B} contains no noise by deriving the essential matrices and bearing vectors with their gradients from the camera poses and reconstructed 3D points.

In other cases, when only pairwise epipolar geometries are known, we propose to apply the following approach using singular value decomposition (SVD).
It is evident that due to $\mat{B}^\trans\matOmega=\mat{0}$, the left-nullspace of \mat{B} is expected to be non-empty. %, containing both columns of \matOmega.
If the null-space is at least two-dimensional, it can contain \matOmega, however the structure of \mat{B} suggests it is three-dimensional.
Thus, we propose to use formula 
%\begin{equation}
    $\matOmega
    =
    \hat{\matOmega} - \mat{U}_{\brc{:,1 \dots 2\brcnorma{\mathcal{V}}-3}}^{} \mat{U}_{\brc{:,1 \dots 2\brcnorma{\mathcal{V}}-3}}^\trans \hat{\matOmega}$,
%\end{equation}
%
where $\mat{U}\mat{S}\mat{V}^\trans=\mat{B}$ is the SVD of \mat{B} and $\mat{U}_{\brc{:,1 \dots 2\brcnorma{\mathcal{V}}-3}}$ is the matrix consisting of the left $2\brcnorma{\mathcal{V}}-3$ columns of \mat{U}.

%%%%%%%%%%%%%%%%%%%%%%%%%%%%%%%%%%%%%%%%%%%%%%%%%%%%%%%%%%%%%%%%%%%%%%%%%%%%%%%%
%%%%%%%%%%%%%%%%%%%%%%%%%%%%%%%%%%%%%%%%%%%%%%%%%%%%%%%%%%%%%%%%%%%%%%%%%%%%%%%%
\paragraph{Refinement of partially affine-covariant regions.}

When a scale- and orientation-covariant detector is applied, \eg AKAZE~\cite{detectors_akaze_alcantarilla_fast_2013} or SIFT~\cite{SIFT2004}, the affine frames can be approximated as $\hat{\mat{M}} \sim \sigma \mat{R}$,
%
%\begin{equation}
%    \hat{\mat{M}} \sim \sigma \mat{R},
%\end{equation}
%
where $\sigma \in \mathbb{R}^\text{+}$ is the scale of the local frame, while $\mat{R} \in \mathbb{R}^{2\times2}$ encodes the dominant orientation of the underlying region.
%The proof that this is the best $L_2$-approximation is submitted as supplementary material. % (TODO)
%
%It is validated by an $L_2$-approximation of a local affine transformation that removes shear and sets a uniform scale: $\hat{\mat{M}} = \mat{U}\mat{S}\mat{V}^\trans \sim \sigma\mat{U}\mat{V}^\trans = \sigma \mat{R}$, where $\mat{R}=\mat{U}\mat{V}^\trans$ and $\sigma = \frac{1}{2}\mathrm{tr}\brc{\mat{S}}$. Thus, $\sigma \mat{R}$ is the closest similarity transformation to $\hat{\mat{M}}$, in the least-squares-sense.
%
Thus, with no special treatment of the partially affine-covariant regions, the proposed method can be applied.% consistent with the camera movement.% using the $\hat{\mat{M}}$ matrices as input similarly as in the previous section. 

%%%%%%%%%%%%%%%%%%%%%%%%%%%%%%%%%%%%%%%%%%%%%%%%%%%%%%%%%%%%%%%%%%%%%%%%%%%%%%%%
%%%%%%%%%%%%%%%%%%%%%%%%%%%%%%%%%%%%%%%%%%%%%%%%%%%%%%%%%%%%%%%%%%%%%%%%%%%%%%%%
%%%%%%%%%%%%%%%%%%%%%%%%%%%%%%%%%%%%%%%%%%%%%%%%%%%%%%%%%%%%%%%%%%%%%%%%%%%%%%%%
\section{Experimental results}

In this section, the proposed method for correcting LAFs is tested both in synthetic experiments and on publicly available real-world datasets. 
First, we show how the proposed method improves the accuracy of detected LAFs. 
Finally, it is demonstrated on a number of real-world problems, \ie homography, surface normal and motion estimation, that using the proposed method leads to superior results. 

%%%%%%%%%%%%%%%%%%%%%%%%%%%%%%%%%%%%%%%%%%%%%%%%%%%%%%%%%%%%%%%%%%%%%%%%%%%%%%%%
%%%%%%%%%%%%%%%%%%%%%%%%%%%%%%%%%%%%%%%%%%%%%%%%%%%%%%%%%%%%%%%%%%%%%%%%%%%%%%%%
\paragraph{Synthetic experiments.}

To test the proposed method in a fully controlled environment, $N$ cameras were generated by their projection matrices looking towards the origin, each located in a random surface point on a sphere of radius $5$. 
Then, a random 3D oriented point, at most one unit away from the origin and with random normal, was projected into the cameras.
The ground truth LAF in each image was calculated from the projection matrix and the surface normal as in~\cite{barath2016novel}.
Zero-mean Gaussian noise with $\sigma$ standard deviation was added to both the point locations and affine parameters. 
Each reported result is averaged over $1,000$ runs. The processing time of 5 views is $\approx0.03$ms.

\begin{figure*}[ht]
	\begin{subfigivan}{0.325\linewidth}
	    \includegraphics[width=1.0\textwidth]{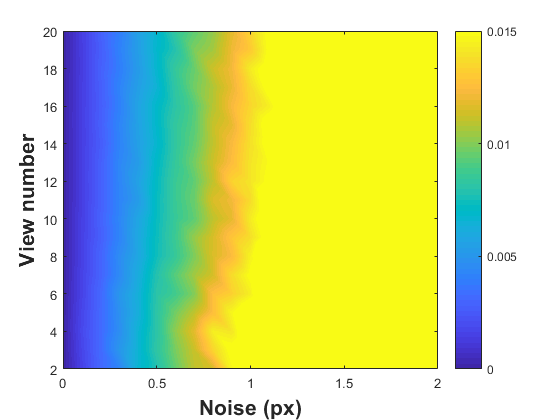}
	    %%\caption{}
        \label{fig:synthetic_a}
	\end{subfigivan}\hfill
	\begin{subfigivan}{0.325\linewidth}
	    \includegraphics[width=1.0\textwidth]{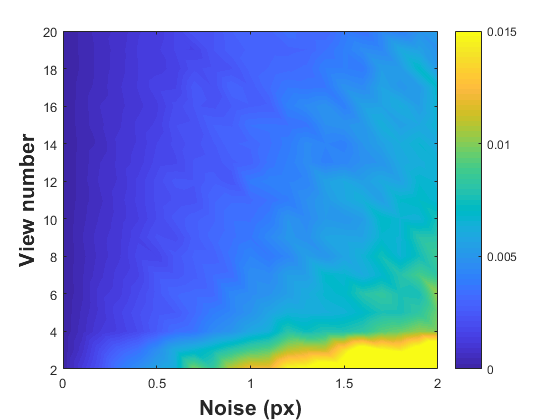}
	    %\caption{}
        \label{fig:synthetic_b}
	\end{subfigivan}\hfill
	\begin{subfigivan}{0.325\linewidth}
	    \includegraphics[width=1.0\textwidth]{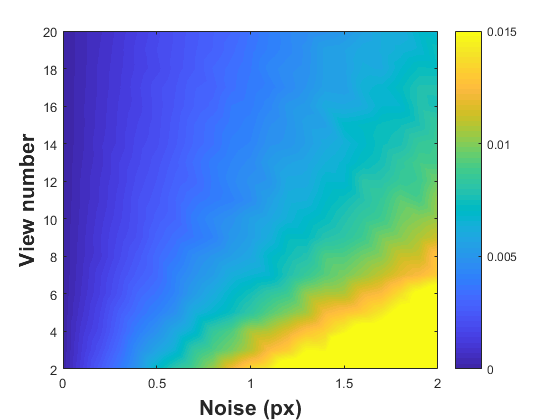}
	    %\caption{}
        \label{fig:synthetic_c}
	\end{subfigivan}
%	\begin{subfigivan}{0.325\linewidth}
%	    \includegraphics[width=1.0\textwidth]{assets/increasing_noise_1.pdf}
%	    \caption{2 views}
%	\end{subfigivan}\hfill
%	\begin{subfigivan}{0.325\linewidth}
%	    \includegraphics[width=1.0\textwidth]{assets/increasing_noise_2.pdf}
%	    \caption{3 views}
%	\end{subfigivan}\hfill
%	\begin{subfigivan}{0.325\linewidth}
%	    \includegraphics[width=1.0\textwidth]{assets/increasing_noise_4.pdf}
%	    \caption{5 views}
%	\end{subfigivan}\\
%	\begin{subfigivan}{0.325\linewidth}
%	    \includegraphics[width=1.0\textwidth]{assets/increasing_noise_9.pdf}
%	    \caption{10 views}
%	\end{subfigivan}\hfill
%	\begin{subfigivan}{0.325\linewidth}
%	    \includegraphics[width=1.0\textwidth]{assets/increasing_noise_14.pdf}
%	    \caption{15 views}
%	\end{subfigivan}\hfill
%	\begin{subfigivan}{0.325\linewidth}
%	    \includegraphics[width=1.0\textwidth]{assets/increasing_noise_19.pdf}
%	    \caption{20 views}
%	\end{subfigivan}
    \caption{\textit{Accuracy of the proposed method.} 
    The error  of the noisy (a) and corrected (b--c) LAFs are plotted as the function of the noise level $\sigma$ (horizontal axis; in pixels) and view number (vertical axis).
    Plot (a) shows the error of the input. 
    For (b), the ground truth fundamental matrix was used. 
    For (c), $\mat{F}$ was estimated from the noisy points. 
    The error is $(1 / K) \sum_{i = 1}^K \norm{\textbf{I} - \textbf{M}_{i,\text{gt}}^{-1} \textbf{M}_{i,\text{est}} }_\text{F}$, where $K$ is the number of views, $\textbf{I}$ is a $3 \times 3$ identity matrix, $\textbf{M}_{i,\text{gt}}$ and $\textbf{M}_{i,\text{est}}$ are, respectively, the ground truth and estimated LAFs in the $i$th view.
    }
\label{fig:example_results}
\end{figure*}

In Fig.~\ref{fig:synthetic_a}, the errors of the noisy LAFs, \ie the input without the correction, are plotted as the function of the noise level $\sigma$ (horizontal axis; in pixels) and view number (vertical). 
In Fig.~\ref{fig:synthetic_b}, the errors of the corrected frames are shown when using the ground truth fundamental matrices for the correction. 
In Fig.~\ref{fig:synthetic_c}, the errors are shown when the $\mat{F}$s are estimated from the noisy point coordinates applying the normalised 8-point algorithm~\cite{hartley2003multiple}. 
It can be seen that the proposed method is consistent, \ie, the more views are given, the more accurate the results are. 
Also, Fig.~\ref{fig:synthetic_c} shows that the method \textit{significantly improves the input LAFs} even if the estimated epipolar geometries are noisy. 
More detailed evaluation and processing times are provided in the supplementary material. 

%%%%%%%%%%%%%%%%%%%%%%%%%%%%%%%%%%%%%%%%%%%%%%%%%%%%%%%%%%%%%%%%%%%%%%%%%%%%%%%%
%%%%%%%%%%%%%%%%%%%%%%%%%%%%%%%%%%%%%%%%%%%%%%%%%%%%%%%%%%%%%%%%%%%%%%%%%%%%%%%%
\paragraph{Comparing feature extractors.}
%\subsection{Real-world evaluation in an SfM environment}

% short description/introduction
In this section, commonly used feature extractors are applied to images of the Strecha dataset~\cite{strecha2008benchmarking} and their outputs are corrected by the proposed method.
The dataset\footnote{Available at \url{http://cvlabwww.epfl.ch/data/multiview/denseMVS.html}} consists of six image sequences of size $3072 \times 2048$ of buildings. Both the intrinsic and extrinsic parameters are given for all images. 
To obtain ground truth LAFs in each image sequence, we first applied an SfM pipeline~\cite{moulon2013global} with the known camera parameters obtaining a number of points along the images. 
Then, the points were manually assigned to dominant planes. 
Since each plane defines a homography between every view pair, the ground truth affine correspondences between the view pairs were calculated from the homography parameters as described in~\cite{barathPRL2017}. 
The evaluated extractors can be divided into three groups: 
(i) scale and rotation-covariant ones, like SIFT~\cite{SIFT2004}, AKAZE~\cite{detectors_akaze_alcantarilla_fast_2013}, Hessian~\cite{detectors_library_vlfeat}, Difference of Gaussians (DoG)~\cite{detectors_library_vlfeat}, and Harris-Laplace (Harris)~\cite{detectors_library_vlfeat}.
(ii) Affine-covariant extractors using the Baumberg-iteration \cite{baumberg_reliable_2000} such as Hessian-Aff, DoG-Aff and Harris-Aff,
and (iii) methods using simulated views, such as ASIFT~\cite{Morel2009}, AAKAZE, etc.
In the experiments, the VlFeat library~\cite{detectors_library_vlfeat} provides the Hessian, DoG and Harris extractors, and their covariant counterparts: Hessian-Aff, DoG-Aff and Harris-Aff using its built-in version of the shape adaptation procedure (\ie, the Baumberg iteration).
We used the SIFT and AKAZE implementations included in OpenMVG~\cite{moulon2016openmvg}.
For AAKAZE and ASIFT, the view-simulation of \cite{Morel2009} is used, feeding warped versions of the input images to the detectors.

% the procedure 
For the experiments, we used a modified version of OpenMVG~\cite{moulon2016openmvg} which, together with the point coordinates, stores the LAFs.
For each detector, we performed feature extraction, then established multi-view correspondences.
The Global SfM pipeline~\cite{moulon2013global} of OpenMVG estimated the camera motion and created a 3D point cloud of the scene.
A robust triangulation procedure then established multi-view tracks of LAFs, with geometrically consistent centroids.
Finally, the corrected LAFs were obtained by the proposed method using the estimated poses.

The results are in Table~\ref{tab:strecha-results}. 
After the header, the odd rows report the accuracy of the extracted LAFs.
The even rows show the quality of the corrected ones. 
Pairs of rows show the results of a particular detector. 
The sequences of the Strecha dataset are from the 3rd to 8th columns. 
The last two columns show the mean and median errors on the entire dataset. It can be seen that the proposed method almost \textit{always improved} the input LAFs.
The most accurate detector is AAKAZE with the proposed correction. Also, it can be seen that the proposed technique significantly improves partially affine-covariant detectors, \eg SIFT, as well. 
We were surprised that SIFT, without the correction, obtains more accurate LAFs than ASIFT on average. The reason is however simple. ASIFT extracts, on average, ten times more correspondences which greatly influences its mean error. However, the median error of ASIFT is 0.19 while that of SIFT is 0.20. Other detectors are in the supplementary material. 

\newcommand\B[1]{\mba{1.2}{#1}}
\newcommand\BR[1]{\mbaa{1.2}{#1}}
\begin{table*}[ht]
\centering
%\resizebox{\textwidth}{!}{%
\begin{tabular}{@{}cc|d{1.2} d{1.2} d{1.2} d{1.2} d{1.2} d{1.2}|d{1.2}d{1.2}}
\toprule
detector & \multicolumn{1}{c|}{LAF type} & \multicolumn{1}{c}{(a)} & \multicolumn{1}{c}{(b)} & \multicolumn{1}{c}{(c)} & \multicolumn{1}{c}{(d)} & \multicolumn{1}{c}{(e)} & \multicolumn{1}{c|}{(f)} & \multicolumn{1}{c}{mean} & \multicolumn{1}{c}{median} \\ \midrule
\multirow{2}{*}{AKAZE} & \multicolumn{1}{c|}{Extracted} & 0.23 & 0.22 & 0.27 & \B{0.27} & 0.30 & 0.26 & 0.26 & 0.20 \\
 & \multicolumn{1}{c|}{Corrected} & \B{0.12} & \B{0.12} & \B{0.14} & 0.62 & \B{0.18} & \BR{0.17} & \B{0.22} & \B{0.09} \\ \midrule
\multirow{2}{*}{SIFT} & \multicolumn{1}{c|}{Extracted} & 0.22 & 0.22 & 0.23 & 0.26 & 0.31 & 0.29 & 0.26 & 0.20 \\
 & \multicolumn{1}{c|}{Corrected} & \B{0.14} & \B{0.12} & \B{0.13} & \B{0.18} & \B{0.18} & \BR{0.21} & \B{0.16} & \B{0.11} \\ \midrule
\multirow{2}{*}{Hessian} & Extracted & 0.25 & 0.25 & 0.26 & 0.26 & 0.33 & 0.29 & 0.27 & 0.22 \\
 & Corrected & \B{0.14} & \B{0.14} & \B{0.12} & \B{0.16} & \B{0.20} & \BR{0.20} & \B{0.16} & \B{0.11} \\ \midrule \midrule
\multirow{2}{*}{Hessian-Aff} & Extracted & 0.29 & 0.29 & 0.29 & 0.37 & 0.41 & 0.35 & 0.33 & 0.25 \\
 & Corrected & \B{0.13} & \B{0.12} & \B{0.13} & \B{0.23} & \B{0.16} & \BR{0.18} & \B{0.16} & \B{0.10} \\ \midrule
\multirow{2}{*}{DoG-Aff} & Extracted & 0.25 & 0.25 & \B{0.29} & 0.27 & 0.43 & 0.39 & 0.31 & 0.19 \\
 & Corrected & \B{0.08} & \B{0.08} & 0.40 & \B{0.16} & \B{0.27} & \BR{0.23} & \B{0.20} & \B{0.07} \\ \midrule \midrule
\multirow{2}{*}{AAKAZE} & Extracted & 0.26 & 0.25 & 0.31 & 0.32 & 0.30 & 0.28 & 0.29 & 0.22 \\
 & Corrected & \B{0.11} & \B{0.10} & \B{0.13} & \B{0.18} & \B{0.12} & \BR{0.13} & \B{0.13} & \B{0.08} \\ \midrule
\multirow{2}{*}{ASIFT} & Extracted & 0.24 & 0.24 & 0.25 & 0.28 & 0.31 & 0.30 & 0.27 & 0.19 \\
 & Corrected & \B{0.11} & \B{0.11} & \B{0.12} & \B{0.17} & \B{0.14} & \BR{0.16} & \B{0.14} & \B{0.08} \\ \bottomrule
\end{tabular}
%}%

\caption{\textit{Comparison of feature detectors} in terms of the accuracy of the obtained LAFs. 
The accuracy (same metric as in Fig.~\ref{fig:example_results}) of the extracted and corrected (by the proposed method) LAFs are put in the odd and even rows, respectively.
The scenes (columns) of the Strecha dataset: (a)~\texttt{castle-P19}, (b)~\texttt{castle-P30}, (c)~\texttt{entry-P10}, (d)~\texttt{fountain-P11}, (e)~\texttt{herz-jesus-P25} and (f)~\texttt{herz-jesus-P8} were fed into the \cite{moulon2016openmvg} SfM pipeline. The proposed method almost \textit{always improve} the extracted LAFs.
}
\label{tab:strecha-results}
\end{table*}

%%%%%%%%%%%%%%%%%%%%%%%%%%%%%%%%%%%%%%%%%%%%%%%%%%%%%%%%%%%%%%%%%%%%%%%%%%%%%%%%
%%%%%%%%%%%%%%%%%%%%%%%%%%%%%%%%%%%%%%%%%%%%%%%%%%%%%%%%%%%%%%%%%%%%%%%%%%%%%%%%
\noindent
\textbf{Application: homography estimation}
using affine correspondences (ACs). 
We used the Strecha dataset and, solely for validation purposes, the manually annotated homographies, similarly as in the previous section. 
Affine correspondences were estimated by the AAKAZE method since it leads to the most accurate LAFs (see Table~\ref{tab:strecha-results}). 
As homography estimator, we chose the HAF method from~\cite{barathPRL2017} which estimates the homography from a single affine correspondence and the fundamental matrix. 

To test the proposed method, we iterated through every possible image pair in each sequence. 
For each pair, the following procedure was applied to every AC:

\noindent
\textbf{1}. The AC is assigned to the closest, in terms of re-projection error, homography $\mat{H}^*$ from the manual annotation. If the error is bigger than $3.0$ px, the AC is rejected.  \\
\noindent
\textbf{2}. Homography $\mat{H}$ is estimated from the AC and fundamental matrix by the HAF method.  \\
\noindent
\textbf{3}. Given the ground truth inliers $\mathcal{I}^*$ of $\mat{H}^*$ from the manual annotation, the proportion of them (\ie, $|\mathcal{I}| / |\mathcal{I}^*|$, where $\mathcal{I} \subseteq \mathcal{I}^*$ and $\forall \pt{p} \in \mathcal{I}$ is inlier of $\mat{H}$) being inlier of $\mat{H}$ as well is measured. The threshold is set to $3.0$ px. \\
\textbf{4}. To measure how a state-of-the-art robust estimator benefits from the proposed method, we applied the local optimisation step of USAC~\cite{raguram2013usac} to $\mat{H}$. 

\begin{figure*}[ht]
%\begin{center}
	\begin{subfigivan}{0.305\linewidth}
	    \includegraphics[width=0.99\textwidth]{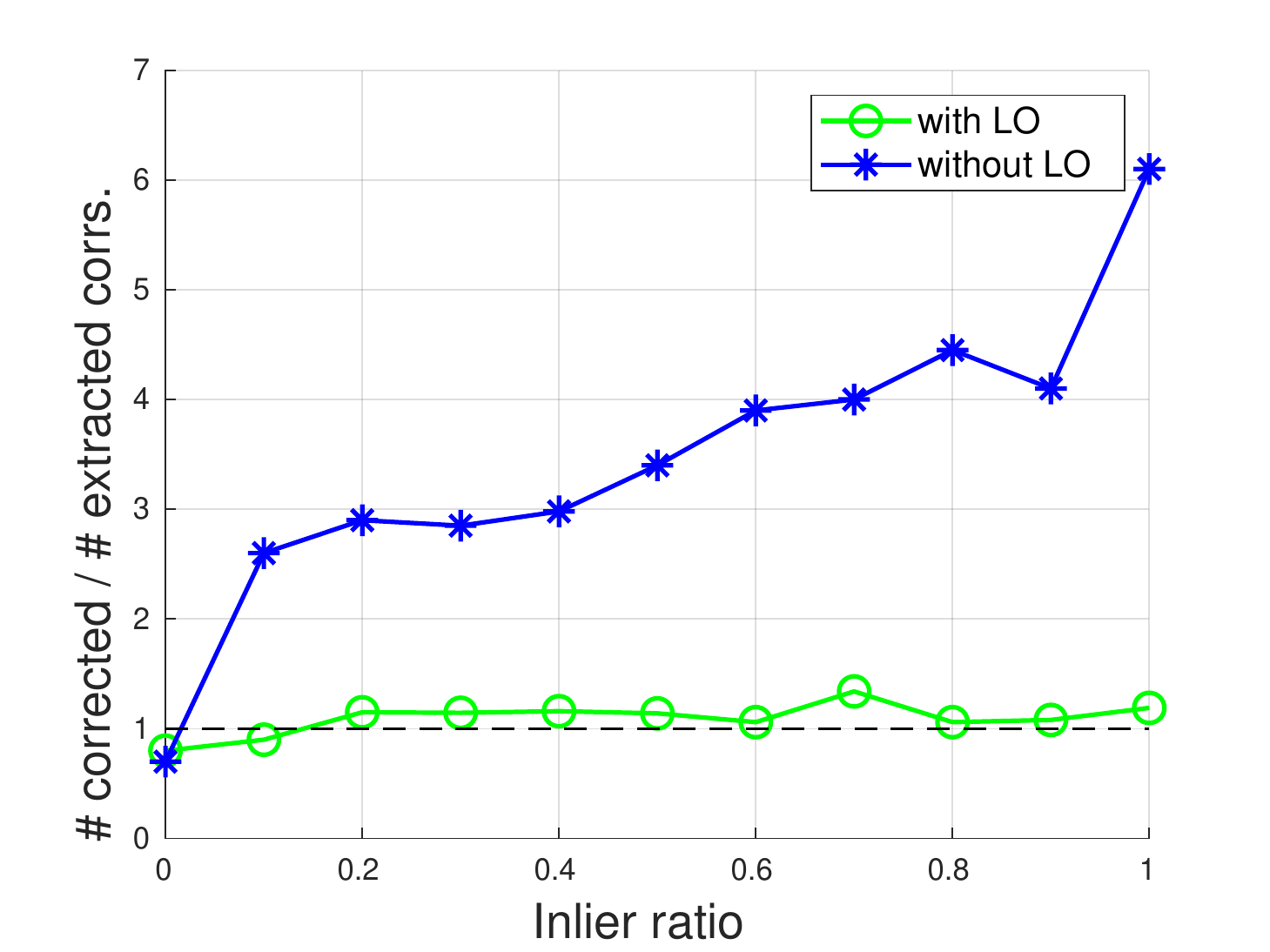}
	    %%\caption{}
        \label{fig:inlier_frequencies}
	\end{subfigivan}
	\begin{subfigivan}{0.325\linewidth}
	    \includegraphics[width=1.0\textwidth]{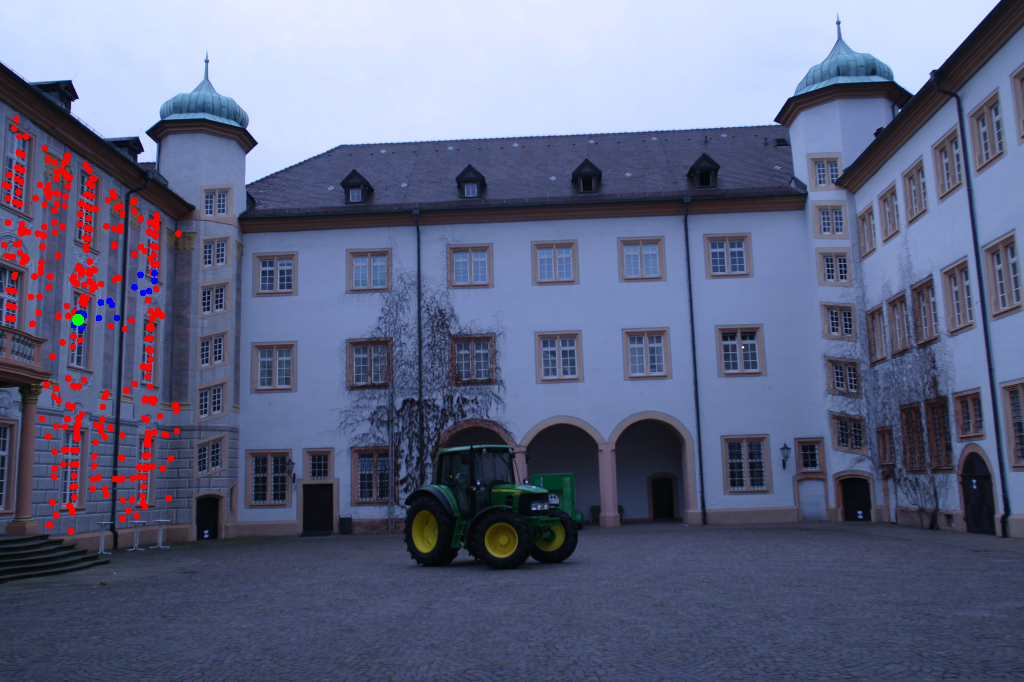}
	    %%\caption{}
        \label{fig:example_image_detected}
	\end{subfigivan}
	\begin{subfigivan}{0.325\linewidth}
	    \includegraphics[width=1.0\textwidth]{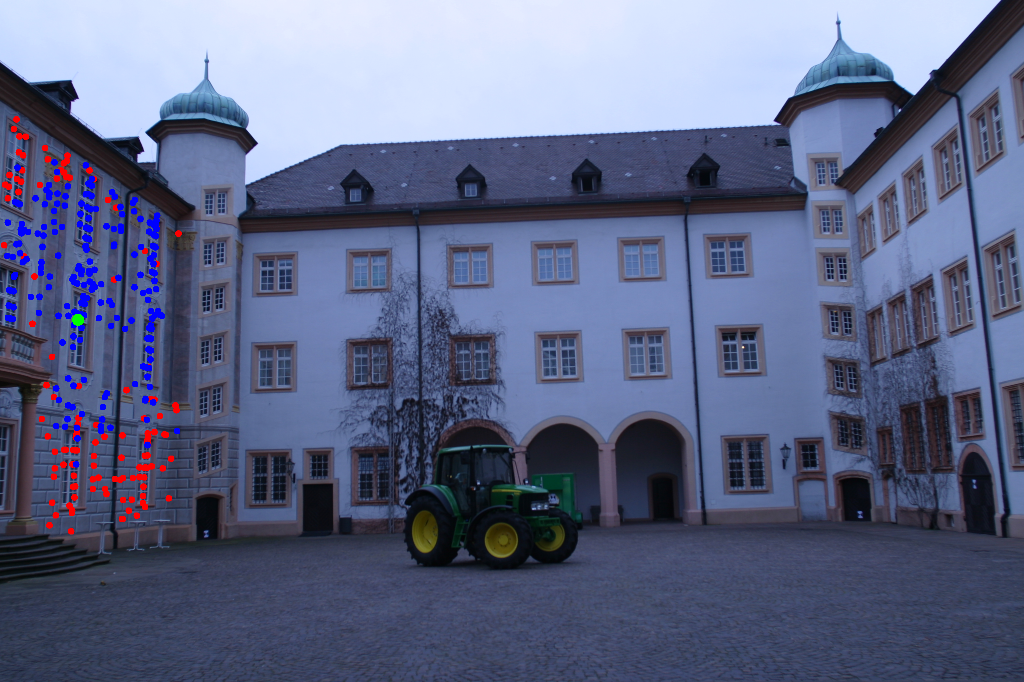}
	    %%\caption{}
        \label{fig:example_image_corrected}
	\end{subfigivan}
	\begin{subfigivan}{0.305\linewidth}
	    \includegraphics[width=0.99\textwidth]{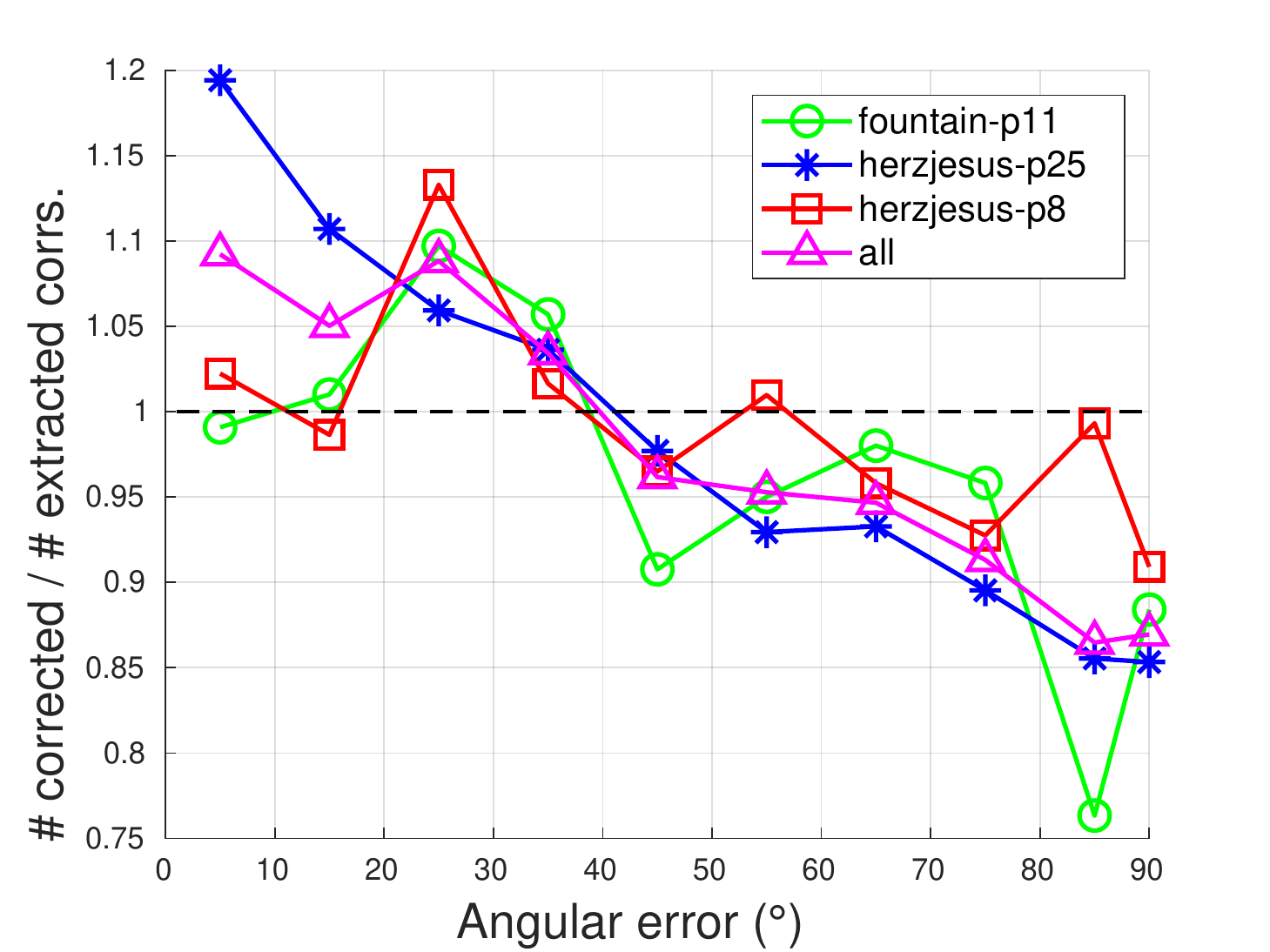}
	    %%\caption{}
        \label{fig:inlier_frequencies_normal}
	\end{subfigivan}
	\begin{subfigivan}{0.345\linewidth}
	    \includegraphics[width=1.1\textwidth]{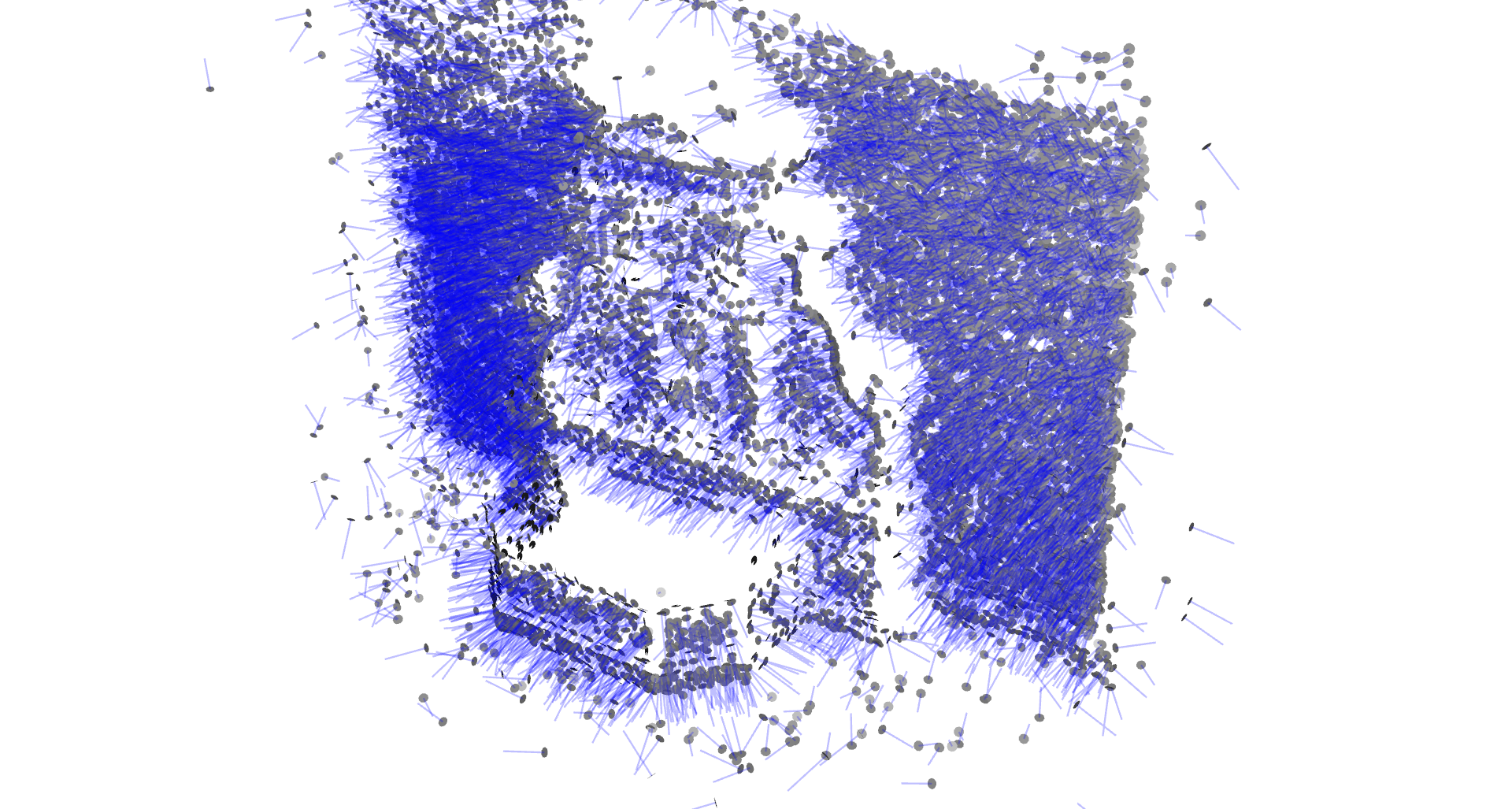}
	   % \includegraphics[width=1.0\textwidth]{assets/fountain2.png}
	    %%\caption{}
        \label{fig:reconstructed_normals}
	\end{subfigivan}
	\begin{subfigivan}{0.325\linewidth}
	    \includegraphics[width=1.0\textwidth]{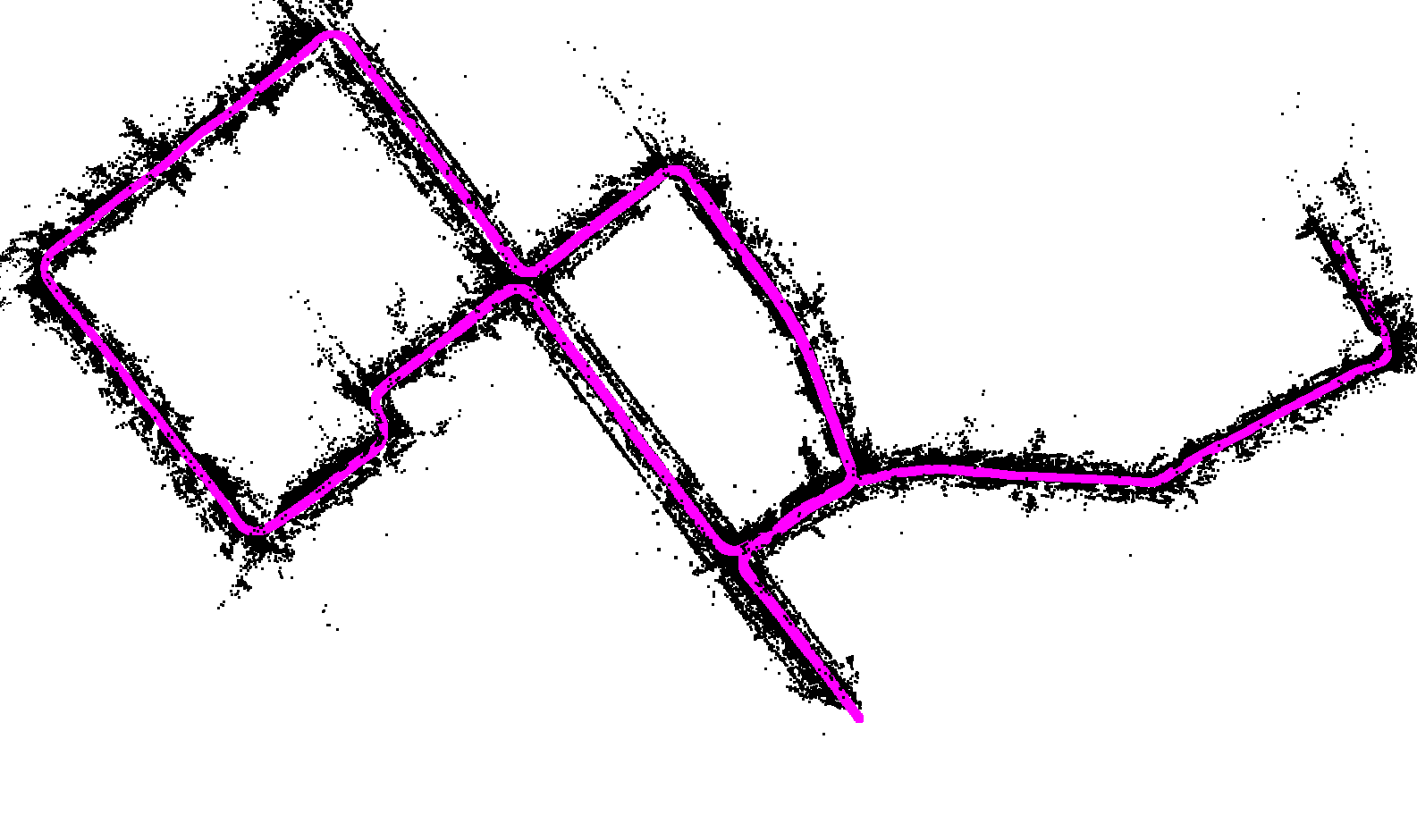}
	    %%\caption{}
        \label{fig:camera_trajectory}
	\end{subfigivan}
%\end{center}
    \caption{\textit{Accuracy of AC-wise homography and surface normal estimation.} 
    \textbf{(a)} Homographies. The avg.\ improvement (vertical axis) of the corrected LAFs compared to the extracted ones are plotted as the function of the inlier ratio (horizontal) with and without local optimisation. 
    We explain the figure via examples. 
    The green curve at $0.4$ inlier ratio is $\approx3$. Thus, there are three times more ACs leading to $0.4$ inlier ratio among the corrected ones than in the extracted set. 
    Accordingly, there are $>6$ times more ACs leading to $\approx1$ inlier ratio. 
    \textbf{(b--c)} Left image of an example image pair from the \texttt{castle-p30} sequence.
    $\mat{H}$ is estimated from the (b) extracted and (c) corrected ACs centred on the green point. 
    The inliers of $\mat{H}$ (blue points) and rest of the points from the same plane (red) are drawn.
    The corrected AC led to ${\approx}12$ times more inliers than the extracted one.
    \textbf{(d)} shows the same for normal estimation as (a) for $\mat{H}$ estimation. \Eg, in the \texttt{herzjesus-p8} scene (blue curve) there are 1.2 times more (vertical axis) corrected ACs leading to 5$^{\circ}$ angular error (horizontal) than in the extracted set. \textbf{(c)} An example scene with reconstructed normals (blue lines) and points. \textbf{(f)} Example scene from the KITTI dataset~\cite{geiger2013vision}.}% showing the reconstructed camera trajectory.}
\end{figure*}

In Fig.~\ref{fig:inlier_frequencies}, the average improvement of the corrected LAFs is plotted as the function of the inlier ratio (horizontal axis) with and without local optimisation. 
We explain the figure through an example. The value of the green curve at $0.4$ inlier ratio is approx.\ $3$. This means that there are three times more ACs amongst the corrected ones than in the extracted correspondence set which led to $0.4$ inlier ratio. Accordingly, there are more than $6$ times more correspondences leading to $\approx1$ inlier ratio. 
Also, the ratio of ACs leading to $0$ inliers is decreased significantly. Originally, $99,331$ extracted ACs led to $\approx0$ inliers and $29,848$ of them were upgraded by the proposed method to have higher inlier ratio. 
This improvement is slightly less significant, although consistent, when the local optimisation is applied.

In conclusion, homography estimation benefits from the corrected ACs significantly. 
The $\mat{H}$s estimated from the corrected ACs are more capable of distinguishing the sought inliers. 

%The improvement is calculated as $|\mathcal{I}^c| / |\mathcal{I}|$, where $\mathcal{I}, \mathcal{I}^c \subseteq \mathcal{I}^*$ are, respectively, the inliers of the homography estimated from the extracted and corrected LAFs.
%We calculated the quality, \ie the number of inliers of the implied homography, of the detected affine correspondence and that of the corrected one. 
%The number of correspondences was then counted which lead to a particular inlier ratio. 
%The ratio of these two numbers are the reported improvement. In total, $203,205$ correspondences were used in the evaluation.
%
%For example, \textit{without local optimisation}, $99,331$ detected correspondences lead to homographies with $0.0$ inlier ratio, and $69,483$ ones after the correction. Therefore, the proposed correction reduced the ratio of entirely useless correspondences by $31\%$. 
%Also, $562$ detected and $1,303$ corrected correspondences lead to $0.5$ inlier ratio.
%On average, the detected correspondences lead to $5.57\%$ inlier ratio, while the corrected ones to $9.07\%$.
%\textit{After local optimisation} on each correspondence, the detected ones lead to $51.57\%$ inlier ratio on average, while the corrected ones to $57.50\%$. 

%%%%%%%%%%%%%%%%%%%%%%%%%%%%%%%%%%%%%%%%%%%%%%%%%%%%%%%%%%%%%%%%%%%%%%%%%%%%%%%%
%%%%%%%%%%%%%%%%%%%%%%%%%%%%%%%%%%%%%%%%%%%%%%%%%%%%%%%%%%%%%%%%%%%%%%%%%%%%%%%%
\noindent
\textbf{Application: surface normal estimation} using affine correspondences.
We applied the multi-view least-squares optimal method from~\cite{barathEichhardtHajderTIP2019} to estimate surface normals from the extracted and corrected LAFs. 
The used sequences from the Strecha dataset are \texttt{fountain} \texttt{-}\texttt{p11}, \texttt{herzjesus-p8} and \texttt{herzjesus-p25} since those are the only ones with publicly available ground truth 3D point cloud. We estimated the ground truth surface normals from the point clouds. 
The error is calculated as the angular error (in degrees) between the reconstructed surface normal and the ground truth one.

Fig.~\ref{fig:inlier_frequencies_normal} shows the improvement (vertical axis), by using the proposed method as a pre-processing step, plotted as the function of the angular error (horizontal). The same property is shown as for homographies in Fig.~\ref{fig:inlier_frequencies}. 
For example, in the \texttt{herzjesus-p8} scene (blue curve), there are 1.2 times more (vertical axis) corrected ACs leading to 5$^{\circ}$ angular error (horizontal) than in the extracted set. 
Also, if all scenes are considered (red curve), there are significantly fewer corrected LAFs leading to $>40^{\circ}$ angular error than in the extracted LAF set. This means that the curve is under 1. In conclusion, the proposed method \textit{improves surface normal estimation} via improving its input significantly.
In Fig.~\ref{fig:reconstructed_normals}, an example scene with reconstructed normals (blue lines) and points are shown. 

\renewcommand\B[1]{\mba{2.2}{#1}}
\renewcommand\BR[1]{\mbaa{2.2}{#1}}

%%%%%%%%%%%%%%%%%%%%%%%%%%%%%%%%%%%%%%%%%%%%%%%%%%%%%%%%%%%%%%%%%%%%%%%%%%%%%%%%
%%%%%%%%%%%%%%%%%%%%%%%%%%%%%%%%%%%%%%%%%%%%%%%%%%%%%%%%%%%%%%%%%%%%%%%%%%%%%%%%
\noindent
\textbf{Application: relative motion estimation of a camera rig}
using affine correspondences. 
We used trajectory "$00$" from the KITTI dataset~\cite{geiger2013vision}. 
Multi-view ACs were established in the frames each consisting of a stereo view pair. 
Each two consecutive stereo pairs were used together simulating a rig of four cameras, and the LAFs were corrected using this rig. 
%The refinement was also performed separately for the next two stereo pair in the same manner.
The relative motion was then estimated between the consecutive four-tuples of images (\ie a frame of the rig) using MSAC~\cite{wang2010generalized} and LO$^\text{+}$-MSAC~\cite{lebeda2012fixing} robust methods. The 2AC solver~\cite{eichhardtChetverikovECCV2018} was used as a minimal solver estimating the essential matrix from two affine correspondences.
The error of the estimated poses was calculated using the high-quality ground truth trajectory provided in the KITTI dataset.
In total, $2020$ four-tuples of images, \ie a frame of the rig, were used in the experiments. 
% We emphasise that only two-view relative motion estimation was performed, and the frames during the refinement of LAFs did not overlap.
% Note that while 5PT is a point-correspondence based approach and is unaffected by the LAF-refinement, the 2AC uses two affine correspondences constructed from matched LAFs to solve the problem of relative motion estimation.
% 2AC uses two affine correspondences constructed from matched LAFs to solve the problem of relative motion estimation. 
% MV-EG-$L_2$ has a definite positive effect on the results of 2AC, when using either of the robust methods.

Table~\ref{tab:camera-rig} reports the accuracy of the robust estimation applied to the extracted and corrected LAFs. 
Due to the improved LAFs, the robust estimation did fewer iterations (3rd column) and, thus, it  sped up (4th). 
Also, the proportion of found inliers is higher (5th), and the \textit{estimated pose is more accurate} if the corrected LAFs were used (6--8th). 
In Fig.~\ref{fig:camera_trajectory}, the ground truth camera trajectory is shown. 

%The runtime and number of iterations of robust estimation is a notable aspect, since approaches based on affine correspondences  claim to improve the most on these aspects compared to point-based ones. We also provide the mean and median rotation and translation (angular) errors, measured against the high-quality ground truth trajectory of the KITTI dataset. The reported translation error is the angle between the ground truth and the estimated vectors of relative translation.
%Through the refined LAFs we were able to improve the results of 2AC.

\newcommand\G[2]{\mba{#1}{#2}}
\renewcommand\B[1]{\mba{1.2}{#1}}
\begin{table}[]
\centering
\resizebox{0.8\textwidth}{!}{%
\begin{tabular}{@{}rr | d{2.0}d{2.2}d{2.2} d{1.2}d{1.2} d{1.2}d{1.2}}
\toprule
\mc{robust method} & \mcc{c|}{LAF type}     & \mc{iters.} & \mc{t (ms)} & \mc{inliers} & \mc{mean $\rho$} & \mc{med.\ $\rho$} & \mc{mean $\tau$} & \mc{med.\ $\tau$} \\ \midrule
\multirow{2}{*}{MSAC} & Extracted & 28 & 5.1 & 59.0\% & 0.66 & 0.18 & 8.11 & 2.62 \\
 & Corrected & \G{2.0}{26} & \G{1.1}{4.5} & \G{2.2}{60.4\%}& \B{0.61} & \B{0.17} & \B{7.31} & \B{2.35} \\ \midrule
\multirow{2}{*}{LO$^+$-MSAC} & Extracted & 23 & 6.5 & 74.8\% & 0.45 & \B{0.09} & 5.00 & 1.30 \\
 & Corrected & \G{2.0}{22} & \G{1.1}{5.7} & \G{2.2}{75.2\%} & \B{0.38} & \B{0.09} & \B{4.18} & \B{1.28} \\ \bottomrule
\end{tabular}
}

\caption{\textit{Relative motion estimation of a camera rig} (from the KITTI dataset~\cite{geiger2013vision}) using the extracted and corrected LAFs. 
MSAC~\cite{wang2010generalized} and LO$^\text{+}$-MSAC~\cite{lebeda2012fixing} were used as robust estimators and 2AC~\cite{eichhardtChetverikovECCV2018} as a minimal solver. 
The reported properties (averaged over $2020$ frames) are: number of iterations (3rd column), runtime (in ms; 4th), proportion of inliers (in $\%$; 5th), rotation ($\rho$; 6--7th) and translation ($\tau$; 8--9th) errors in degrees.}
\label{tab:camera-rig}
\end{table}

\section{Conclusions}

A closed-form solution is proposed, optimal in the least-squares sense, for correcting the parameters of multi-view affine correspondences represented as a set of LAFs.
The technique requires the epipolar geometry to be pre-estimated between each pair of views and makes the extracted LAFs consistent with the camera movement. 
It is validated both in synthetic experiments and on publicly available real-world datasets that the method almost always improves the input LAFs. 
As a by-product, a number of affine-covariant detectors are compared. On the used datasets, AKAZE with the view synthesizer of \cite{Morel2009} leads to the most accurate LAFs. 
Also, it is shown that it makes the affine frames built on the output of partially affine-covariant detectors, \eg SIFT, significantly more accurate.
As potential applications, it is shown that the proposed correction improves homography, surface normal and relative motion estimation via improving the input of these methods. 
When affine frames are used, we see no reason for not applying the proposed technique. 

{\small
\bibliography{references}
}

\end{document}